\documentclass[11pt]{article}

\usepackage[preprint]{acl}

\usepackage{times}
\usepackage{latexsym}

\usepackage[T1]{fontenc}
\usepackage[utf8]{inputenc}

\usepackage{microtype}

\usepackage{inconsolata}

\usepackage{graphicx}

\usepackage{xspace}
\usepackage{graphicx}
\usepackage{amsmath}
\usepackage{amssymb}
\usepackage{mathtools}
\usepackage{amsthm}
\usepackage{bbm}
\usepackage{enumitem}
\usepackage{multirow}
\usepackage{booktabs}
\usepackage{makecell}
\usepackage[table]{xcolor}
\usepackage{amsmath}
\usepackage[most]{tcolorbox}
\usepackage{fontawesome}
\usepackage[normalem]{ulem}
\usepackage{xspace}
\usepackage{pifont}
\usepackage{amssymb}
\usepackage{bbding}
\usepackage{pifont}
\usepackage{makecell} 
\usepackage{flushend}
\usepackage{multirow}
\usepackage{wrapfig}
\usepackage{enumitem}
\usepackage{adjustbox}
\usepackage{makecell}
\usepackage{colortbl}
\usepackage{pgf}
\usepackage{bm}
\usepackage{amsmath}

\usepackage[normalem]{ulem}
\usepackage{wrapfig} 
\usepackage{xcolor}
\usepackage{pifont}
\usepackage{listings}
\usepackage{cleveref} 
\usepackage{algorithm}
\usepackage{algpseudocode}

\usepackage{cuted}
\usepackage{graphicx, booktabs, multirow, array, amsmath, amssymb, listings, pifont, natbib}   
\usepackage[table]{xcolor}     
\usepackage{booktabs, siunitx, xcolor, colortbl}
\useunder{\uline}{\ul}{}
\newcommand{\method}{\textsc{PromptSD}\xspace}

\definecolor{backcolour}{rgb}{0.95,0.95,0.92}  
\definecolor{codegreen}{rgb}{0,0.6,0}
\definecolor{codegray}{rgb}{0.5,0.5,0.5}
\definecolor{codepurple}{rgb}{0.58,0,0.82}

\title{One Student, Many Teachers: Multi-Task On-Policy Distillation via Soft-Prompt Privileged Context}

\author{%
  Yingzi Ma\textsuperscript{1}\thanks{Equal contribution. Corresponding to \texttt{yma382@wisc.edu, chaoweixiao@jhu.edu}.} \quad
  Zichen Zhu\textsuperscript{3}\footnotemark[1] \quad
  Ming Jiang\textsuperscript{1} \quad
  Chaowei Xiao\textsuperscript{2} \\[0.6em]
  \normalsize
  \textsuperscript{1}University of Wisconsin--Madison \quad
  \textsuperscript{2}Johns Hopkins University \\
  \textsuperscript{3}Nanyang Technological University
}

\begin{document}
\maketitle
\begin{abstract}
On-policy self-distillation (OPSD) teaches large language models new skills through a teacher that shares the student's backbone and supervises its own rollouts. Existing teachers either inject privileged context at the input---inducing post-hoc rationalization---or fine-tune weights, accumulating drift and forgetting across tasks. We propose \method, whose teacher differs from the student only by a learnable soft prompt: trained on $(x, y_\text{gold})$ pairs with the backbone frozen, the prompt yields a task-specific teacher that preserves the student's exact representational geometry. \method\ extends naturally to multi-task settings by routing each example in a merged corpus to its corresponding soft-prompt teacher, allowing a single student to absorb knowledge from $K$ teachers in parallel; at inference, all prompts are discarded. On Qwen3-1.7B-Base and Phi-4-mini-instruct across four tasks (Science, Tool Use, Biology, Math), the single-task variant (OPD with a PT teacher) matches or exceeds full fine-tuning while training orders of magnitude fewer parameters, and the multi-task variant achieves the best overall average ($56.2$ on Qwen3-1.7B-Base) while preserving general-capability benchmarks---in contrast to sequential SFT, which degrades both.
\end{abstract}

\section{Introduction}

\begin{figure*}[!t]
    \centering
    \includegraphics[width=\linewidth]{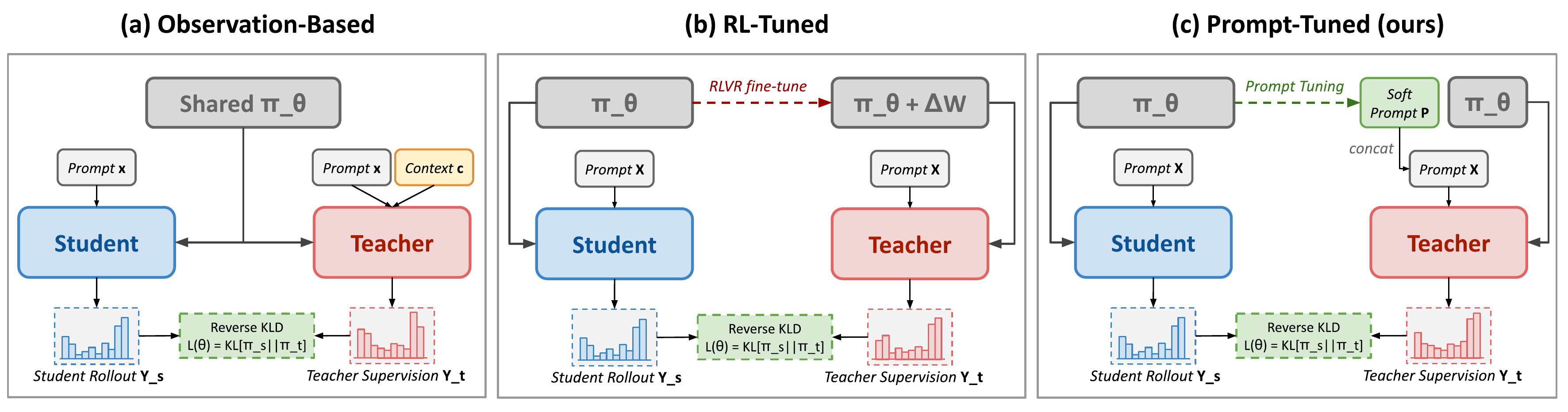}
     \vspace{-1em}
    \caption{\textbf{Three families of privileged-information teachers for on-policy distillation.} All three share the same on-policy reverse-KL objective $\mathcal{L}(\theta) = \mathrm{KL}[\pi_s \,\|\, \pi_t]$ over student rollouts $Y_s$, but differ in how the teacher acquires privileged information. (a) \emph{Observation-based} (SDFT, OPSD, SDPO): teacher and student share weights $\pi_\theta$; the teacher is conditioned on extra input-side context $c$. (b) \emph{RL-tuned} (CoPD): the teacher is fine-tuned with RLVR, drifting to $\pi_\theta + \Delta W$. (c) \emph{Prompt-tuned (ours)}: the teacher shares the student's weights but prepends a learnable soft prompt $P$ obtained via prompt tuning; only $P$ differs between teacher and student.}
    \label{fig:overview}
    \vspace{-1em}
\end{figure*}

On-policy distillation (OPD) has emerged as a core post-training technique for large language models~\citep{yang2025qwen3,xiao2026mimo,zeng2026glm}: student rollouts paired with per-token teacher distributions provide a denser supervision signal than off-policy SFT~\citep{arora2022exposure,agarwal2024policy}. Recent work realizes the teacher $\pi_\text{T}$ in four structurally distinct ways: \emph{observation-based} self-teachers inject privileged context $c$ at the input---demonstrations~\citep{shenfeld2026self}, gold answers~\citep{zhao2026self}, or verifier-filtered rollouts~\citep{hubotter2026reinforcement}; \emph{RL-tuned} teachers fine-tune the base with verifiable rewards~\citep{gu2026co,li2026rethinking}; \emph{cross-size} teachers distill from a larger same-family model~\citep{yang2025qwen3,agarwal2024policy}; and \emph{SFT-tuned} teachers fine-tune on $(x, y_\text{gold})$ pairs~\citep{thinkingmachines2025opd}. Amid this rapid proliferation, a basic question is rarely asked: \emph{which teacher should be paired with which task, and once we have a good teacher, why distill from it rather than deploy it directly?} \looseness=-1

Each family answers this only partially. \citet{li2026rethinking} formalize two conditions for effective OPD: \emph{thinking-pattern consistency} on student-visited states, and a \emph{non-trivial capability gap}. While observation-based teachers (OPSD) naturally preserve the first by sharing weights with the student, they face a structural risk we call \emph{post-hoc rationalization}~\citep{turpin2023language,arcuschin2025chain}: conditioning the teacher on the gold answer leads the teacher to justify rather than derive its trace, a failure mode previously identified in CoT distillation as the ``reasoning shortcut between question and gold answer''~\citep{wang2023scott}. RL-based teachers face a different problem: they require verifiable rewards (unavailable for many tasks), and recent analysis~\citep{chen2026does} argues that RL primarily amplifies existing capabilities rather than injecting new ones---a limitation visible in domains like vision-language-action modeling, where SFT cold-start is a prerequisite for RL~\citep{wang2025alpamayo}. We therefore pursue SFT-based teacher construction. \looseness=-1

SFT methods span a spectrum of distributional perturbation: full fine-tuning (FFT), LoRA~\citep{hu2022lora}, adapters~\citep{houlsby2019parameter}, and prompt tuning (PT)~\citep{lester2021power}. PT occupies a uniquely favorable position for OPD: it trains a small set of continuous embeddings prepended to the input while leaving all transformer weights frozen. As illustrated in Figure~\ref{fig:overview}, this places PT (panel c) at a distinct point in the design space relative to observation-based (panel a) and RL-tuned (panel b) teachers: the soft prompt $P$ acts as a parametric form of observation, expressed in continuous embedding space rather than discrete tokens, while keeping the transformer weights identical to the student's. Against weight-modifying methods (FFT, LoRA), PT preserves thinking-pattern consistency and avoids the weight drift and ``intruder dimensions'' that compound across sequential tasks~\citep{biderman2024lora,shuttleworth2026lora}. Against observation-based teachers, PT's soft prompt does not condition the teacher on any answer, eliminating post-hoc rationalization. This design also bounds catastrophic forgetting~\citep{kirkpatrick2017overcoming,luo2025empirical}: by keeping the teacher distributionally close to the base, PT minimizes the per-step KL signal driving student weight updates. \looseness=-1

In light of these observations, we propose \method, an on-policy distillation framework that uses a prompt-tuned teacher and extends to multiple tasks via parallel routing. \method has two components. First, the \emph{prompt-tuned teacher}: for each task $k$, we attach a set of learnable continuous embeddings $P_k$ to the frozen student $\pi_\theta$ and optimize $P_k$ on $(x, y_\text{gold})$ via standard supervised fine-tuning, yielding a task-specific teacher $\pi_{\theta + P_k^\star}$ that shares all transformer parameters with the student and differs only in input-side conditioning. Second, \emph{parallel multi-task distillation}: rather than training one student per task or chaining tasks sequentially, we merge all task datasets into a single corpus where each example retains a task tag $k$, and at each student update we route the example to its corresponding teacher $\pi_{\theta + P_k^\star}$. The student is updated via on-policy reverse-KL on its own rollouts, with the teacher's supervision bounded by the prompt-induced gap from the base. At deployment we discard all soft prompts and return $\pi_\theta$ alone. \looseness=-1

We validate \method through two sets of experiments on Qwen3-1.7B-Base and Phi-4-mini-instruct. First, on single-task distillation across four tasks (Science, Tooluse, Biology, Math), single-task OPD with a PT teacher matches or exceeds SFT-tuned teachers (FFT, LoRA), and is the only configuration whose average accuracy exceeds its teacher's accuracy on a target task. Second, multi-task \method\ achieves the best overall average ($\mathbf{56.2}$ on Qwen3-1.7B-Base, $+2.3$ over the strongest single-task baseline) while simultaneously preserving general-capability benchmarks (MMLU-Pro~\cite{wang2024mmlu}, HellaSwag~\cite{zellers2019hellaswag}, TruthfulQA~\cite{lin2022truthfulqa})---in contrast to sequential SFT, which loses both trained-task accuracy and general capability to catastrophic forgetting. Our analysis further yields two principles for OPD teacher design: teacher \emph{family} should match the task type (SFT for capability injection, RL for elicitation), and teacher \emph{scale} matters less than whether the teacher carries new knowledge. \looseness=-1

\section{Related Work}

\subsection{On-Policy Distillation}
On-policy distillation (OPD) addresses the train--test mismatch of off-policy knowledge distillation by training the student on its own rollouts under teacher per-token supervision. GKD~\citep{agarwal2024policy} interpolates between off-policy and on-policy mixtures and supports multiple divergence choices, while MiniLLM~\citep{gu2024minillm} minimizes reverse KL to prevent the student from overestimating the teacher's low-probability tails. OPD has since been adopted in industrial post-training pipelines (Qwen3~\citep{yang2025qwen3}, MiMo~\citep{xiao2026mimo}, GLM-5~\citep{zeng2026glm}) as a competitive complement to SFT and outcome-reward RL. A subsequent line of \emph{self-distillation} variants couples OPD with a single backbone serving as both teacher and student, where the teacher gains its informational advantage from privileged context: SDFT~\citep{shenfeld2026self} prepends in-context demonstrations, OPSD~\citep{zhao2026self} prepends gold reasoning traces, and SDPO~\citep{hubotter2026reinforcement} uses verifier-filtered student rollouts as context. CoPD~\citep{gu2026co} departs from this input-side family by RLVR-tuning the teacher's weights. Mechanism analyses by \citet{li2026rethinking} identify thinking-pattern consistency and a non-trivial capability gap as the two governing conditions of successful OPD---a framework on which we directly build (\S\ref{sec:motivation}). None of these methods construct the teacher via prompt tuning, and the parameter-side overlay design we propose is, to our knowledge, the first to satisfy both conditions by construction. \looseness=-1

\subsection{Post-Training Techniques}
\begin{figure*}[!t]
    \centering
    \includegraphics[width=\linewidth]{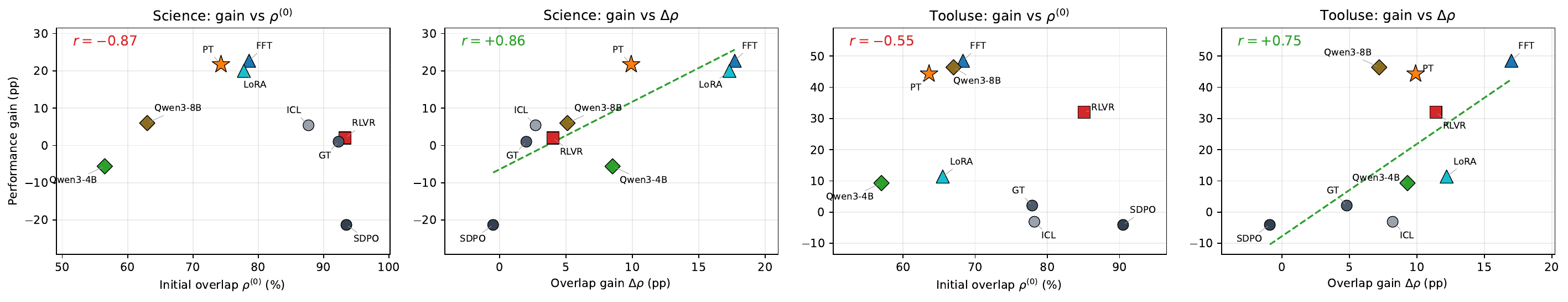}
     \vspace{-2em}
    \caption{\textbf{Performance gain vs.\ overlap dynamics across teacher families.} For each teacher (markers), we plot the 1.7B-Base student's task accuracy gain (pp over zero-shot) against (left of each pair) the initial overlap $\rho^{(0)}$ and (right of each pair) the overlap gain $\Delta\rho = \rho^{(S)} - \rho^{(0)}$, on Science and Tooluse. Dashed lines indicate the linear fit. \emph{Initial overlap alone is not predictive}: high-$\rho^{(0)}$ teachers like \textsc{RLVR} and \textsc{SDPO} produce near-zero or negative gains, while moderate-$\rho^{(0)}$ teachers like \textsc{PT}, \textsc{FFT}, and \textsc{LoRA} produce the largest gains. \emph{Overlap gain $\Delta\rho$ is strongly predictive of student gain} (right panels): teachers occupying the moderate-overlap regime (D1) and producing substantial $\Delta\rho$ (D2) consistently yield the best students. }
    
\label{fig:overlap_gain}
    \vspace{-1em}
\end{figure*}

Supervised fine-tuning (SFT)~\citep{ouyang2022training} provides dense token-level supervision from $(x, y_\text{gold})$ pairs but trains exclusively on expert trajectories, suffering from \emph{exposure bias} at inference~\citep{arora2022exposure} and inducing substantial distributional drift that manifests as catastrophic forgetting~\citep{kirkpatrick2017overcoming,luo2025empirical}. Parameter-efficient variants such as LoRA~\citep{hu2022lora} mitigate but do not eliminate this drift, with recent analyses identifying ``intruder dimension'' artifacts that concentrate update mass far from the base manifold~\citep{shuttleworth2026lora,biderman2024lora}. Reinforcement learning approaches train on student-generated trajectories: RLHF~\citep{ouyang2022training} learns a reward model from human preferences and optimizes it with PPO, while RLVR~\citep{guo2025deepseek,shao2024deepseekmath} replaces the reward model with programmatic verifiers. Both face structural limits in our setting: reward sparsity yields high-variance trajectory-level gradients, and recent analyses argue that RLVR primarily amplifies skills already latent in the base rather than injecting new ones~\citep{chen2026does}---consistent with the stagnant $\Delta\rho$ we observe for RLVR teachers in Figure~\ref{fig:overlap_gain}. Direct Preference Optimization~\citep{rafailov2023direct} sidesteps reward modeling but is restricted to preference pairs and has been shown to induce implicit reward models with limited out-of-distribution generalization. OPD is complementary to all three: it inherits SFT's dense token-level signal, RL's on-policy training distribution, and avoids verifier and preference-annotation requirements. Our work asks a sharper question---which teacher construction best preserves the conditions under which dense on-policy supervision actually works.\looseness=-1

\section{Motivation}
\label{sec:motivation}

\subsection{Desiderata for a Suitable OPD Teacher}
\label{subsec:desiderata}

Building on \citet{li2026rethinking} and our reading of OPD training dynamics, we identify three desiderata an OPD teacher $\pi_\text{T}$ should jointly satisfy with respect to a student $\pi_s = \pi_\theta$. Let $t$ index the token position within a student rollout $Y_s$ on prompt $x$, and let
\begin{align}
S^{(p)}_t &= \mathrm{TopK}\!\left(\pi_s(\cdot \mid x, Y_s^{<t}),\, k\right), \notag \\
S^{(q)}_t &= \mathrm{TopK}\!\left(\pi_\text{T}(\cdot \mid x, Y_s^{<t}),\, k\right)
\end{align}
denote the student's and teacher's top-$k$ sets at position $t$. The overlap ratio averages the intersection size over rollout positions and task prompts:
\begin{align}
\rho \;=\; \mathbb{E}_{x,\,Y_s \sim \pi_\theta}\!\left[\, \frac{1}{|Y_s|} \sum_{t=1}^{|Y_s|} \frac{|S^{(p)}_t \cap S^{(q)}_t|}{k} \,\right].
\label{eq:overlap}
\end{align}
We write $\rho^{(s)}$ for the overlap at OPD training step $s$, with $\rho^{(0)}$ the value at initialization and $\rho^{(S)}$ at the end of training. We report $k{=}10$ throughout (results for $k\!\in\!\{1, 5, 50\}$ in the appendix are qualitatively similar).

\noindent\textbf{(D1) Thinking-pattern consistency.} The teacher's top-$k$ distribution on student-visited states must overlap with $\pi_\theta$ enough that the reverse-KL signal is absorbable, but not so much that no learning signal remains. We call this a \emph{moderate-overlap regime}: when $\rho^{(0)}$ is too low, the gap falls outside the student's absorbable range; when $\rho^{(0)}$ is too high, $\pi_\text{T} \approx \pi_\theta$ and the per-token KL gradient vanishes. Formally,
\begin{align}
\rho_{\min} \;<\; \rho^{(0)} \;<\; \rho_{\max}.
\label{eq:d1}
\end{align}

\noindent\textbf{(D2) Capability gap absorbed by the student.} Beyond a moderate starting point, the teacher must encode task-relevant knowledge that $\pi_\theta$ progressively absorbs. We measure this through the \emph{change} in overlap rather than benchmark accuracy: a real capability gap manifests as steadily rising $\rho^{(s)}$, while a teacher that merely amplifies existing behaviors~\citep{chen2026does} leaves overlap stagnant despite higher accuracy. We require
\begin{align}
\Delta := \rho^{(S)} - \rho^{(0)} \;>\; \delta.
\label{eq:d2}
\end{align}

\noindent\textbf{(D3) No post-hoc rationalization.} The teacher's conditioning context at distillation time should ideally not contain $y_\text{gold}$, as doing so leads the teacher to justify rather than derive its reasoning---a pattern the student inherits without access to the answer at inference~\citep{wang2023scott}. This is particularly acute when $y_\text{gold}$ is only a short final answer: the teacher has no intermediate trace to follow and fabricates a derivation that lands on the predetermined answer, biasing token-level distributions toward backward-chained justification. \looseness=-1

\subsection{Existing Teacher Families}
\label{subsec:preliminary}

We examine how six teacher families fare against D1--D3 by plotting each teacher's student gain against its overlap dynamics ($\rho^{(0)}$ and $\Delta\rho$) on Science and Tooluse (Figure~\ref{fig:overlap_gain}).

\noindent\textbf{Observation-based teachers} share weights with the student (D1 satisfied) but place $y_\text{gold}$ in the teacher's input, violating D3---as a gold answer~\citep{zhao2026self}, demonstrations as in \textsc{SDFT}~\citep{shenfeld2026self}, or verifier-filtered rollouts~\citep{hubotter2026reinforcement}. \textsc{ICL} yields \emph{negative} $\Delta\rho$ on Science: its logits depend on demonstration context the student never sees, compounded by SLMs' weak in-context learning. \textsc{GT} and \textsc{SDPO} occupy the high-$\rho^{(0)}$ end of Figure~\ref{fig:overlap_gain} ($\ge 85$\%) yet produce near-zero student gain---the gold-conditioned teacher is already nearly identical to the student on student-visited prefixes.

\noindent\textbf{RL-tuned teachers} satisfy D3 but require task-specific verifiers and primarily amplify existing behaviors rather than inject new ones~\citep{chen2026does}. \textsc{RLVR} sits in the high-$\rho^{(0)}$ region with stagnant $\Delta\rho$: the teacher's distribution is already nearly fully covered by the student, so the reverse-KL signal vanishes and D2 fails. \looseness=-1

\noindent\textbf{Large-sized teachers} satisfy D2 and D3 but suffer cross-model parameterization mismatch---distinct pretraining trajectories and tokenization-level idiosyncrasies place the teacher's distribution off the student's behavioral manifold. \textsc{Qwen3-4B} and \textsc{Qwen3-8B} populate the left edge of the $\rho^{(0)}$ plots ($\approx 55$--$65$\%) and produce small student gains, consistent with D1's lower bound being violated. \looseness=-1

\noindent\textbf{SFT-tuned teachers and the SFT trap.} \textsc{FFT} and \textsc{LoRA} satisfy D1--D3 and occupy the upper-right corner of the $\Delta\rho$ plots---largest $\Delta\rho$ and largest student gains. Yet the desiderata miss a \emph{side effect}: the same $\Delta W$ that enables capability absorption (D2) also perturbs the teacher's distribution on \emph{out-of-task} inputs (\textsc{FFT} broadly; \textsc{LoRA} via low-rank ``intruder dimensions''~\citep{shuttleworth2026lora}), which the student inherits as catastrophic forgetting in sequential settings (\S\ref{sec:experiments_continual}). Prompt tuning sits naturally between observation-based teachers (privileged context $c$ at the input, D3-violating) and SFT-tuned teachers ($\Delta W$-driven drift): the soft prompt $P$ provides a \emph{learnable, parametric} privileged context while keeping $\Delta W{=}0$ in the transformer body. The teacher $\pi_{\theta+P}$ thus satisfies D1--D3 without off-manifold drift. In Figure~\ref{fig:overlap_gain}, \textsc{PT} lands in the moderate-$\rho^{(0)}$ region with the largest student gain---the only same-base teacher achieving both D1 and D2 jointly. \looseness=-1

\section{\method}
\label{sec:method}

\method realizes the teacher design motivated in \S\ref{subsec:desiderata}: a soft-prompt overlay on the frozen student backbone, supervising the student via on-policy reverse-KL distillation. We describe the three components in turn: constructing the soft-prompt teacher (\S\ref{subsec:build_teacher}), the on-policy distillation objective (\S\ref{subsec:opd_objective}), and the parallel multi-task extension (\S\ref{subsec:multitask}).

\subsection{Building a Soft-Prompt Teacher}
\label{subsec:build_teacher}

Given a base model $\pi_\theta$ and a task dataset $\mathcal{D} = \{(x_i, y_i)\}_{i=1}^N$, we construct a teacher by attaching $L$ continuous embeddings $P = [p_1, \ldots, p_L] \in \mathbb{R}^{L \times d}$ to the input embedding layer of $\pi_\theta$, where $d$ is the hidden dimension. For input $x$ with embedding sequence $E(x)$, the teacher's forward pass treats $P$ as a virtual prefix:
\begin{align}
\pi_{\theta+P}(\cdot \mid x) \;=\; \pi_\theta\!\left(\cdot \mid [P; E(x)]\right),
\label{eq:pt_teacher}
\end{align}
where $[\,\cdot\,;\,\cdot\,]$ denotes concatenation along the sequence axis. Only $P$ is trainable; all transformer weights $\theta$ remain frozen. We optimize $P$ by minimizing the next-token cross-entropy on $\mathcal{D}$:
\begin{align}
\mathcal{L}_{\text{PT}}(P) \;=\; -\,\mathbb{E}_{(x, y) \sim \mathcal{D}} \!\left[ \log \pi_{\theta+P}(y \mid x) \right],
\label{eq:pt_loss}
\end{align}
yielding $P^\star$ that encodes task knowledge in $L \times d$ parameters---typically $10$--$100\times$ smaller than LoRA at standard rank settings---while leaving $\theta$ untouched.

\subsection{On-Policy Soft-Prompt Distillation}
\label{subsec:opd_objective}

With $\pi_{\theta + P^\star}$ fixed, we distill its capability into $\pi_\theta$ via on-policy reverse-KL distillation. At each step, we sample $x \sim \mathcal{D}$, generate a student rollout $Y_s \sim \pi_\theta(\cdot \mid x)$, and evaluate both models on each prefix $Y_s^{<t}$. Let
\begin{align}
p_t(\cdot) &\;\triangleq\; \pi_\theta(\cdot \mid x, Y_s^{<t}), \notag \\
q_t(\cdot) &\;\triangleq\; \pi_{\theta + P^\star}(\cdot \mid x, Y_s^{<t})
\label{eq:pq_def}
\end{align}
denote the per-step student and teacher distributions. The student is updated to minimize
\begin{align}
\mathcal{L}_{\text{OPD}}(\theta) \;=\; \mathbb{E}_{x,\,Y_s \sim \pi_\theta} \!\left[ \sum_{t=1}^{|Y_s|} \mathrm{KL}(p_t \,\|\, q_t) \right].
\label{eq:opd_objective}
\end{align}
Crucially, both models receive the same prompt $x$---$P^\star$ acts as a parameter-side overlay on the teacher rather than an input-side context shift---so the teacher's per-token supervision is computed on exactly the states the student visits. At inference, $P^\star$ is discarded and the deployed model is $\pi_\theta$ alone, incurring no parameter or compute overhead relative to the base.

\begin{figure}[!t]
    \centering
    \includegraphics[width=\linewidth]{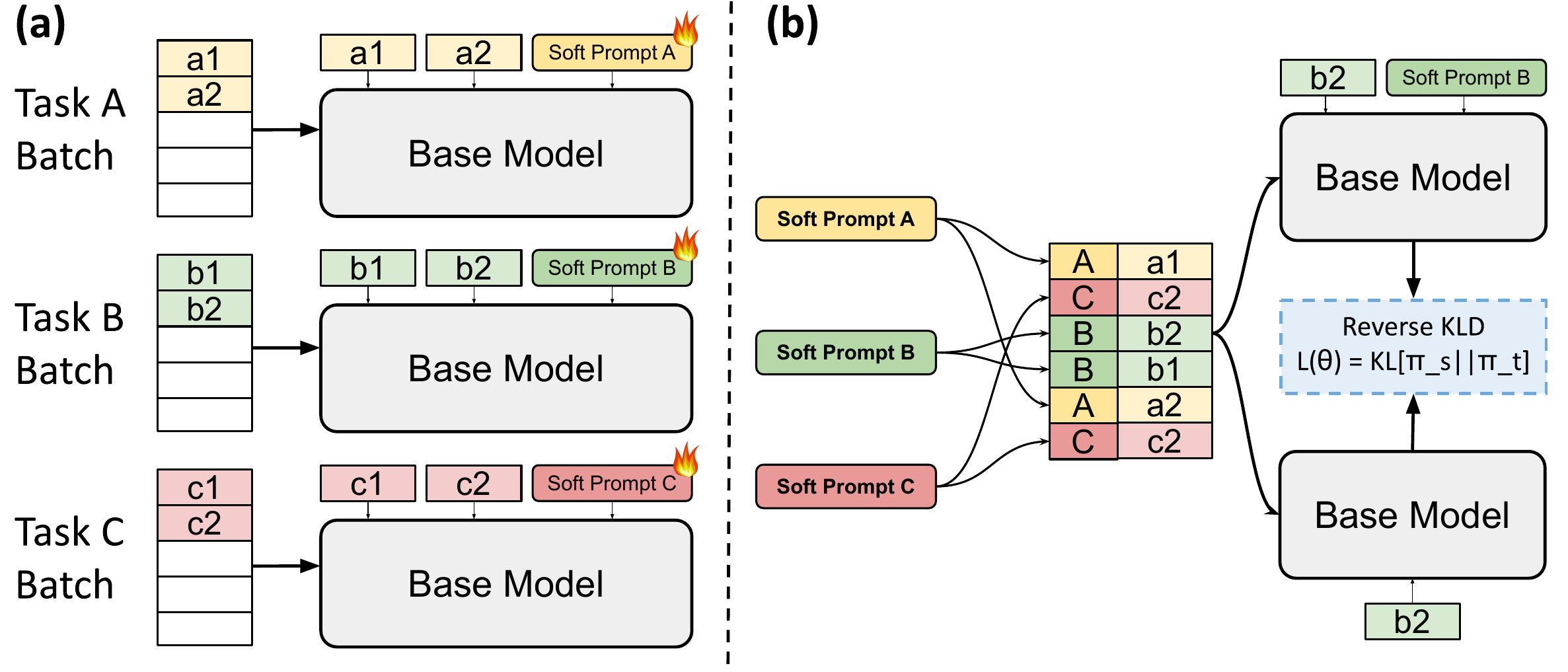}
     \vspace{-1em}
    \caption{\textbf{Overview of \method.} \textbf{(a)} Task-specific soft prompts $\{P_1^\star, \ldots, P_K^\star\}$ are trained in parallel on $\{\mathcal{D}_1, \ldots, \mathcal{D}_K\}$ via next-token cross-entropy (Eq.~\ref{eq:pt_loss}); the base model is frozen. \textbf{(b)} Datasets are merged with task tags; each example is routed to its soft-prompt teacher $\pi_{\theta + P_k^\star}$, while the student $\pi_\theta$ rolls out on the input alone. The student is updated via per-token reverse KL on student-visited states (Eq.~\ref{eq:multitask}). At inference, soft prompts are discarded.}
\label{fig:method_overview}
    \vspace{-1em}
\end{figure}

\subsection{Parallel Multi-Task Distillation}
\label{subsec:multitask}

\method extends naturally to settings with $K$ tasks $\{\mathcal{T}_1, \ldots, \mathcal{T}_K\}$, each with dataset $\mathcal{D}_k$. We first train $K$ task-specific soft prompts $\{P_1^\star, \ldots, P_K^\star\}$ in parallel by applying Eq.~\ref{eq:pt_loss} independently on each $\mathcal{D}_k$; the prompts share no parameters and incur no inter-task interference. We then form a merged distillation corpus
\begin{align}
\mathcal{D}_{\text{merged}} = \bigcup_{k=1}^K \{(x, y, k) : (x, y) \in \mathcal{D}_k\},
\label{eq:merged_corpus}
\end{align}
in which each example retains a task tag $k$ identifying its source. At each distillation step, we sample $(x, y, k) \sim \mathcal{D}_{\text{merged}}$, generate $Y_s \sim \pi_\theta(\cdot \mid x)$, and update the student against the task-routed teacher $\pi_{\theta + P_k^\star}$:
\begin{align}
\mathcal{L}_{\text{OPD}}^{\text{multi}}(\theta) \;=\; \mathbb{E}_{(x, y, k),\, Y_s} \!\left[ \sum_{t} \mathrm{KL}(p_t \,\|\, q_t^{(k)}) \right],
\label{eq:multitask}
\end{align}
where $q_t^{(k)}(\cdot) = \pi_{\theta + P_k^\star}(\cdot \mid x, Y_s^{<t})$. Task-specific knowledge is routed through the corresponding soft prompt at supervision time but unified into shared parameters at the student. Because all $K$ prompts are simultaneously available throughout training, this scheme is order-independent and avoids the per-task forgetting incurred by sequential continual learning---a property we examine empirically in \S\ref{sec:exp}. \looseness=-1

\begin{table*}[!t]
\centering
\tiny
\caption{Per-method results on four independently fine-tuned tasks (\textbf{Qwen3-1.7B-Base}; no sequential continual learning). \textbf{Acc} = task accuracy (\%). \textbf{$\Delta\rho$} measures capability transfer: $\rho^{(S)}-\rho^{(0)}$ in percentage points, where $\rho$ is the mean top-10 next-token overlap between student and teacher over teacher-forced validation rollouts. Positive $\Delta\rho$ means the student moved \emph{toward} the teacher (genuine absorption rather than amplifying prior behavior). Best and second-best \textbf{Acc} per column among same-size students are \textbf{bolded} and \underline{underlined}. $^*$ denotes multi-task learning.}
\label{tab:per_task_teacher_comp_17b}
    \renewcommand{\arraystretch}{1.05}
\resizebox{\linewidth}{!}{
\begin{tabular}{lcccccccccc}
\toprule
                                \multirow{2.5}{*}{\textbf{Method}}  & \multicolumn{2}{c}{\textbf{Science}} & \multicolumn{2}{c}{\textbf{Tooluse}} & \multicolumn{2}{c}{\textbf{Biology}} & \multicolumn{2}{c}{\textbf{Math}} & \multicolumn{2}{c}{\textbf{Average}} \\
\cmidrule(lr){2-3}\cmidrule(lr){4-5}\cmidrule(lr){6-7}\cmidrule(lr){8-9}\cmidrule(lr){10-11}
 & Acc        & $\Delta\rho$        & Acc        & $\Delta\rho$        & Acc        & $\Delta\rho$        & Acc      & $\Delta\rho$       & Acc        & $\Delta\rho$         \\ \midrule
\rowcolor{gray!8} \multicolumn{11}{c}{\textit{Base model}}                                                                                                                                                                                                                                                                                                                                                                                                                  \\
Zero-Shot      & 29.3  & --  & 7.2   & --  & 32.0  & --  & 16.4  & --  & 21.2  & -- \\ %\hline

\rowcolor{gray!8} \multicolumn{11}{c}{\textit{Teacher upper bounds}}    \\
{\color[HTML]{7F7F7F} ICL teacher}    & {\color[HTML]{7F7F7F} 35.3}  & {\color[HTML]{7F7F7F} --}  & {\color[HTML]{7F7F7F} 0.0}   & {\color[HTML]{7F7F7F} --}  & {\color[HTML]{7F7F7F} 38.0}  & {\color[HTML]{7F7F7F} --}  & {\color[HTML]{7F7F7F} 57.0}  & {\color[HTML]{7F7F7F} --}  & {\color[HTML]{7F7F7F} 32.6}  & {\color[HTML]{7F7F7F} --} \\
{\color[HTML]{7F7F7F} RLVR teacher}   & {\color[HTML]{7F7F7F} 30.3}  & {\color[HTML]{7F7F7F} --}  & {\color[HTML]{7F7F7F} 40.2}  & {\color[HTML]{7F7F7F} --}  & {\color[HTML]{7F7F7F} 32.0}  & {\color[HTML]{7F7F7F} --}  & {\color[HTML]{7F7F7F} 44.2}  & {\color[HTML]{7F7F7F} --}  & {\color[HTML]{7F7F7F} 36.7}  & {\color[HTML]{7F7F7F} --} \\
{\color[HTML]{7F7F7F} LoRA teacher}   & {\color[HTML]{7F7F7F} 46.7}  & {\color[HTML]{7F7F7F} --}  & {\color[HTML]{7F7F7F} 51.5}  & {\color[HTML]{7F7F7F} --}  & {\color[HTML]{7F7F7F} 26.0}  & {\color[HTML]{7F7F7F} --}  & {\color[HTML]{7F7F7F} 52.8}  & {\color[HTML]{7F7F7F} --}  & {\color[HTML]{7F7F7F} 44.3}  & {\color[HTML]{7F7F7F} --} \\
{\color[HTML]{7F7F7F} FFT teacher}    & {\color[HTML]{7F7F7F} 54.3}  & {\color[HTML]{7F7F7F} --}  & {\color[HTML]{7F7F7F} 51.5}  & {\color[HTML]{7F7F7F} --}  & {\color[HTML]{7F7F7F} 42.0}  & {\color[HTML]{7F7F7F} --}  & {\color[HTML]{7F7F7F} 48.8}  & {\color[HTML]{7F7F7F} --}  & {\color[HTML]{7F7F7F} 49.2}  & {\color[HTML]{7F7F7F} --} \\
{\color[HTML]{7F7F7F} PT teacher}     & {\color[HTML]{7F7F7F} 52.7}  & {\color[HTML]{7F7F7F} --}  & {\color[HTML]{7F7F7F} 47.4}  & {\color[HTML]{7F7F7F} --}  & {\color[HTML]{7F7F7F} 58.0}  & {\color[HTML]{7F7F7F} --}  & {\color[HTML]{7F7F7F} 48.6}  & {\color[HTML]{7F7F7F} --}  & {\color[HTML]{7F7F7F} 50.4}  & {\color[HTML]{7F7F7F} --} \\ %\hline

\rowcolor{gray!8} \multicolumn{11}{c}{\textit{Large-sized cross-size baselines}}                                         \\
OPD (Qwen-4B)  & 23.7  & +8.5  & 16.5  & +9.3  & 20.0  & +0.5  & 86.4  & +4.3  & 36.7  & +5.7 \\
OPD (Qwen-8B)  & 35.3  & +5.1  & 53.6  & +7.2  & 20.0  & +8.3  & 85.2  & +1.3  & 48.5  & +5.5 \\

\rowcolor{gray!8} \multicolumn{11}{c}{\textit{Observation-based baselines}}                                                                                                                                                                                                                                                                                                                                                                                                  \\
SDFT (Demo)              & 34.7                         & +2.7                         & 4.1                          & +8.2                         & 30.0                         & $-$1.2                       & {\ul 67.0}                   & +0.6                         & 34.0                         & +2.6                         \\
OPSD (GT)                & 30.3                         & +2.0                         & 9.3                          & +4.8                         & 30.0                         & +6.8                         & 20.6                         & +0.3                         & 22.6                         & +3.5                         \\
SDPO (Rollout)           & 8.0                          & $-$0.5                       & 3.1                          & $-$0.9                       & 32.0                         & +3.9                         & 16.4                         & $-$3.2                       & 14.9                         & $-$0.2                       \\ %\hline
\rowcolor{gray!8} \multicolumn{11}{c}{\textit{RL-tuned baselines}}                                                                                                                                                                                                                                                                                                                                                                                                            \\
OPD (RLVR)               & 31.3                         & +4.0                         & 39.2                         & +11.4                        & {\ul 42.0}                   & +8.6                         & 41.0                         & +2.9                         & 38.4                         & +6.7                         \\ %\hline
\rowcolor{gray!8} \multicolumn{11}{c}{\textit{SFT-tuned baselines}}                                                                                                                                                                                                                                                                                                                                                                                                           \\
OPD (LoRA)               & 49.3                         & +17.3                        & 18.6                         & +12.2                        & 28.0                         & +30.1                        & 55.4                         & +13.0                        & 37.8                         & +18.2                        \\
OPD (FFT)                & {\ul 52.0}                   & +17.7                        & \textbf{55.7}                & +17.0                        & 38.0                         & +29.6                        & 47.0                         & +13.1                        & 48.2                         & +19.4                        \\
OPD (PT)                 & 51.0                         & +9.9                         & {\ul 51.5}                   & +9.9                         & \textbf{62.0}                & +12.0                        & 51.0                         & +8.1                         & {\ul 53.9}                   & +10.0                        \\

%\hline
\rowcolor{blue!8} \method$^*$ & \textbf{55.3} & +8.8 & 48.4 & +12.9 & {\ul 54.0} & +4.2 & \textbf{67.2} & +8.1 & \textbf{56.2} & +8.5 \\
\bottomrule
\end{tabular}
}
\vspace{-3mm}
\end{table*}

\section{Experiment} \label{sec:exp}

\subsection{Experimental Setup} \label{sec:exp_setup}

\noindent\textbf{Models.}
We use \textbf{Qwen3-1.7B-Base}~\citep{yang2025qwen3} (pretrained, no instruction tuning) as the student backbone throughout, so that all teacher-induced gains are isolated from prior RLHF biases. The same backbone is used to construct every same-size teacher: \textsc{PT} prepends $L$ learnable virtual-token embeddings to the input embedding sequence (\S\ref{subsec:build_teacher}); \textsc{LoRA} uses rank $r{=}16$ on $\{q, k, v, o\}$ projections; \textsc{FFT} updates all transformer weights; \textsc{RLVR} fine-tunes the teacher with GRPO~\citep{shao2024deepseekmath} using each task's verifier as the reward signal; \textsc{ICL} provides $K{=}3$ in-context demonstrations sampled from the train split; \textsc{GT} wraps the gold answer in a fixed prefix template. For cross-size baselines we use Qwen3-4B and Qwen3-8B, both of which share the Qwen3 tokenizer with the 1.7B student so that token-level logits are directly comparable. \looseness=-1

\begin{table*}[!h]
\centering
\footnotesize
\renewcommand{\arraystretch}{0.9}
\caption{\textbf{Trained-task performance vs.\ general-capability retention} (Qwen3-1.7B-Base). \textbf{Avg-T} averages accuracy (\%) over the four target tasks; \textbf{Avg-U} averages over three held-out 5-shot utility benchmarks.}
\vspace{-1em}
\label{tab:method_vs_utility_17b}
\resizebox{\linewidth}{!}{
\begin{tabular}{l|cccc|c|ccc|c}
\toprule
\textbf{Method} & Science & Tooluse & Biology & Math & \textbf{Avg-T} & MMLU-Pro & HellaSwag & TruthfulQA & \textbf{Avg-U} \\
\midrule
{\color[HTML]{7F7F7F} Base} & {\color[HTML]{7F7F7F} 29.3} & {\color[HTML]{7F7F7F} 7.2} & {\color[HTML]{7F7F7F} 32.0} & {\color[HTML]{7F7F7F} 16.4} & {\color[HTML]{7F7F7F} 21.2} & {\color[HTML]{7F7F7F} 35.8} & {\color[HTML]{7F7F7F} 65.5} & {\color[HTML]{7F7F7F} 50.0} & {\color[HTML]{7F7F7F} 50.4} \\
\midrule
SFT (sequential)            & 45.7 & \textbf{58.8} & 44.0 & 29.0 & 44.4 & 29.7 & 62.5 & 46.4 & 46.2 \\
OPD~(PT,~sequential)        & 49.3 & 45.4 & \textbf{60.0} & 53.0 & 51.9 & 34.5 & 66.9 & 52.6 & 51.3 \\
SFT~(multi-task)            & 51.3 & 53.6 & 38.0 & 32.0 & 43.7 & 34.8 & 65.5 & 52.7 & 51.0 \\
OPD~(LoRA,~multi-task)      & 51.7 & 51.5 & 34.0 & 53.0 & 47.5 & 35.6 & 62.8 & \textbf{56.8} & \textbf{51.7} \\
\rowcolor{blue!8}
\method\ (multi-task)       & \textbf{55.3} & 48.4 & 54.0 & \textbf{67.2} & \textbf{56.2} & \textbf{36.6} & \textbf{68.1} & 50.4 & \textbf{51.7} \\
\bottomrule
\end{tabular}}
\vspace{-1.6em}
\end{table*}

\noindent\textbf{Datasets and Metrics.}
We evaluate on four tasks spanning complementary capabilities: \textbf{Science} (4-way MCQ with reasoning tags~\cite{shenfeld2026self}, $n{=}300$), \textbf{Tooluse} (ReAct against Stable-Diffusion tools~\citep{patil2024gorilla}, $n{=}97$), \textbf{Biology} (SciKnowEval~\citep{feng2024sciknoweval}, $n{=}50$), and \textbf{Math} (MATH-500~\citep{hendrycks2021measuring}, $n{=}500$). Each task has its own train/val/test split; soft prompts and on-policy rollouts use the training split, validation drives early stopping, and all reported numbers are on test. We report task-specific accuracy with greedy decoding via vLLM~\citep{kwon2023efficient}. Further details in the appendix.

\noindent\textbf{Implementation.}
Phase A (soft-prompt training): $2000$ steps, AdamW lr $5{\times}10^{-3}$, cosine schedule, effective batch size $16$; only the $L\times d$ prompt embeddings are updated. Phase B (on-policy distillation): $600$ steps, AdamW lr $1{\times}10^{-5}$, with the reverse-KL objective in Eq.~\ref{eq:opd_objective}; the rollout buffer is refreshed every $64$ steps via vLLM. For the multi-task variant (\S\ref{subsec:multitask}), the merged corpus shuffles all four tasks, and each example is routed to its task-specific frozen teacher. All experiments run on $4$ NVIDIA H100 80GB GPUs; a single-task pipeline takes $1$--$2$ GPU-hours, full main results $\sim$$30$ GPU-hours. Further details in the appendix.

\subsection{Main Results}

\begin{figure}[t]
    \centering
    \includegraphics[width=0.9\linewidth]{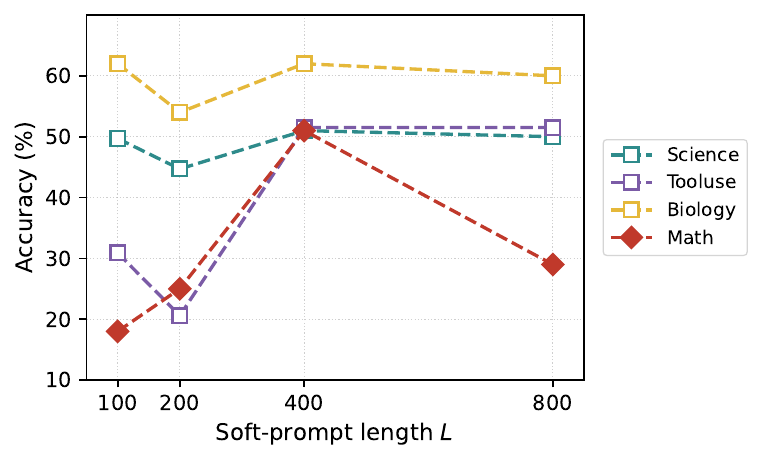}
     \vspace{-1em}
    \caption{\textbf{Soft-prompt length $L$} (number of learnable virtual tokens in the PT teacher), across all four tasks.}
\label{fig:abl_length}
    \vspace{-1.5em}
\end{figure}

\noindent\textbf{Single-Task Distillation.}
Table~\ref{tab:per_task_teacher_comp_17b} compares all teacher families on Qwen3-1.7B-Base. \emph{The teacher upper bounds reveal a task-type contrast: SFT-tuned teachers excel when new knowledge must be injected, while observation- and RL-based teachers compete when capability is already latent and only needs elicitation.} On Science, Tooluse, and Biology, SFT-tuned teachers (PT $50.4$, FFT $49.2$) clearly outperform RLVR ($36.7$) and ICL ($32.6$); on Math, the gap narrows---ICL leads ($57.0$) and SFT teachers fall to $48$--$53$. Among same-size students, \textsc{OPD~(PT)} achieves the best average ($\mathbf{53.9}$) and reaches $\mathbf{62.0}$\% on Biology---$+4$ above its own teacher and $+34$ above \textsc{OPD~(LoRA)}---as the PT teacher injects capability without the off-manifold drift that caps SFT-tuned students. \textsc{OPD~(FFT)} trails PT on average ($48.2$) and its $\Delta\rho{\approx}{+}18$ on every task signals heavy teacher alignment, foreshadowing the catastrophic forgetting reported in \S\ref{sec:experiments_continual}. \emph{Notably, what matters is not teacher scale but whether it carries new knowledge}: cross-size \textsc{OPD~(Qwen-8B)} dominates only on Math ($85.2$) where Qwen3-1.7B-Base is limited, but trails \textsc{OPD~(PT)} on the other three tasks. \looseness=-1

\noindent\textbf{Multi-Task \method.}
\method extends \textsc{OPD~(PT)} by distilling four task-specific PT teachers into a \emph{single} student via merged-corpus routing (\S\ref{subsec:multitask}). \method$^*$ achieves the best average ($\mathbf{56.2}$), exceeding the strongest single-task baseline by $+2.3$ and cross-size \textsc{OPD~(Qwen-8B)} by $+7.7$. The gain is most pronounced on Math ($67.2$ vs.\ $51.0$, $+16.2$ over single-task PT) and Science ($55.3$, the table's best). Multi-task exposure thus provides positive cross-task transfer: jointly training on Science, Tooluse, and Biology rollouts reinforces the reasoning and instruction-following Math latently requires, letting the student elicit capability beyond any single-task teacher. PT teachers compose cleanly---four parametric privileged contexts absorbed by one student without interference. \looseness=-1

\subsection{Continual Learning}\label{sec:experiments_continual}
Per-task SFT (or RLVR) suffices when adapting to a single domain in isolation---SFT injects new knowledge for capability-heavy tasks like Science or Biology, and RLVR sharpens reasoning where the capability is already latent. Realistic deployment, however, is continual: a model must absorb new domains over time without erasing prior capabilities. Table~\ref{tab:method_vs_utility_17b} reports trained-task accuracy (\textbf{Avg-T}) and held-out utility (\textbf{Avg-U}, over MMLU-Pro, HellaSwag, TruthfulQA) across sequential and multi-task baselines.
Sequential SFT fails on both fronts---$44.4$ \textbf{Avg-T} and a drop from the base's $50.4$ to $46.2$ \textbf{Avg-U} (MMLU-Pro $-6.1$)---the SFT trap of \S\ref{subsec:preliminary} realized at the student level: cumulative $\Delta W$ across tasks shifts the student off-manifold on unrelated inputs. Sequential OPD~(PT) already mitigates this ($51.9$/$51.3$) because the PT teacher's bounded per-task perturbation (D1) leaves general capability intact, justifying OPD over deploying the teacher directly. Switching to a multi-task schedule alone does not close the gap: multi-task SFT and OPD~(LoRA) recover Avg-U ($51.0$, $51.7$) but stall at $43.7$ and $47.5$ Avg-T, indicating that the Avg-T bottleneck is the adaptation mechanism, not the schedule. \method\ improves both axes simultaneously ($\mathbf{56.2}$ \textbf{Avg-T}, $\mathbf{51.7}$ \textbf{Avg-U}), with the largest gain on Math ($67.2$ vs.\ sequential PT $53.0$, sequential SFT $29.0$), where cross-task PT exposure unlocks latent capability. The simultaneous Avg-U lift confirms that parametric privileged contexts compose without cumulative drift---\method\ delivers task specialization \emph{and} preserves general capability.

\subsection{Ablation Study}

\noindent\textbf{Soft Prompt Length.}
Figure~\ref{fig:abl_length} ablates the number of learnable virtual tokens $L$ in the PT teacher across all four tasks. No single $L$ dominates: $L{=}400$ is best (or tied for best) on Science, Tooluse, and Math, while Biology shows a U-shaped response peaking at both $L{=}100$ and $L{=}400$ ($62.0$) with a dip at $L{=}200$. The general pattern is that more learnable parameters help capacity-demanding tasks: Math gains $+13$ points from $L{=}100$ to $L{=}400$, and Tooluse jumps from $30.9$ to $51.5$ once $L$ reaches $400$. Beyond $L{=}400$, gains flatten or invert (Math $51.0 \to 29.0$ at $L{=}800$), suggesting diminishing returns. We adopt $L{=}400$ as the default in all main-results experiments as a balance between expressiveness and stability.

\noindent\textbf{Divergence Choice.}
Table~\ref{tab:abl_kl} compares four divergences in the OPD loss with a PT teacher: forward KL (mean-seeking), reverse KL (mode-seeking)~\citep{gu2024minillm}, symmetric KL~\citep{wu2025rethinking}, and Jensen--Shannon~\citep{agarwal2024policy}. On Science, Tooluse, and Biology, all four fall within $\le$$6$ points of each other---divergence choice is largely immaterial for capability-heavy tasks. The exception is Math, where reverse-KL substantially outperforms the alternatives ($51.0$ vs.\ $13$--$18$ for the others), consistent with its mode-seeking behavior concentrating mass on the teacher's high-probability solution tokens rather than averaging across competing trajectories. We adopt reverse-KL as the default for this Math advantage and for consistency with the OPSD literature.

\begin{table}[!t]
\centering
\footnotesize

\caption{\textbf{Divergence choice} in the on-policy distillation loss (PT teacher).
We compare reverse-KL (mode-seeking), forward-KL (mean-seeking), symmetric KL, and
Jensen--Shannon (JSD).}
\vspace{-1em}
\label{tab:abl_kl}
\resizebox{\columnwidth}{!}{
\begin{tabular}{l|cccc}
\toprule
\textbf{Divergence} & \textbf{Science} & \textbf{Tooluse} & \textbf{Biology} & \textbf{Math} \\
\midrule
Forward  & 50.0 & 48.4 & 62.0 & 13.0 \\
Symmetric                & 49.0 & 49.5 & \textbf{64.0} & 18.0 \\
JSD          & 47.0 &  \textbf{51.5} & 58.0 & 14.0 \\
\rowcolor{blue!8}
Reverse   & \textbf{51.0} & \textbf{51.5} & 60.0 & \textbf{51.0} \\
\bottomrule
\end{tabular}}
\vspace{-1em}
\end{table}

\noindent\textbf{Different Backbones.}
To confirm that our findings generalize beyond Qwen3-1.7B-Base, Table~\ref{tab:per_task_phi4mini} repeats the main comparison on \textbf{Phi-4-mini-instruct}~\citep{abdin2024phi}, and three observations carry over. First, \method achieves the best overall average ($\mathbf{56.4}$), \emph{matching the FFT teacher upper bound itself} ($56.4$) while outperforming all same-size single-task baselines including \textsc{OPD~(FFT)} ($55.3$) and \textsc{OPD~(LoRA)} ($51.5$). Second, the task-type contrast persists: SFT-tuned teachers dominate capability-heavy tasks (FFT teacher reaches $52.0$ on Biology vs.\ RLVR's $36.0$), while observation- and RL-based teachers lead on Math, where the base is already strong ($62.0$ for \textsc{OPSD~(GT)} and \textsc{OPD~(RLVR)}, vs.\ FFT teacher's $60.0$). Third, the SFT trap remains visible: \textsc{OPD~(FFT)} falls short of its teacher on three of four tasks (Science, Tooluse, Biology), capped at the teacher's ceiling. By contrast, \method inherits no such cap and reaches the teacher's overall accuracy. Notably, the single-task PT teacher is the relatively weaker teacher among these on this backbone (avg $51.3$), yet \method's multi-task framework still recovers the strongest student ($56.4$)---showing that the merged-corpus routing of \S\ref{subsec:multitask} extracts transferable signal even when individual teachers are imperfect. \looseness=-1
\begin{table}[!t]
\centering
\footnotesize
\caption{Per-method results on four independently fine-tuned tasks based on \textbf{Phi-4-mini-instruct}.}
\vspace{-1em}
\label{tab:per_task_phi4mini}
    \renewcommand{\arraystretch}{1.05}
\resizebox{\linewidth}{!}{
\begin{tabular}{lccccc}
\toprule
\textbf{Method} & \textbf{Science} & \textbf{Tooluse} & \textbf{Biology} & \textbf{Math} & \textbf{Average} \\ \midrule
\rowcolor{gray!8} \multicolumn{6}{c}{\textit{Base model}} \\

Zero-Shot     &  33.7  & 46.4  & 22.0  &55.2  &  39.3 \\ 

\rowcolor{gray!8} \multicolumn{6}{c}{\textit{Teacher upper bounds}} \\
{\color[HTML]{7F7F7F} RLVR teacher } & {\color[HTML]{7F7F7F} 39.3} & {\color[HTML]{7F7F7F} 46.4} & {\color[HTML]{7F7F7F} 36.0} & {\color[HTML]{7F7F7F} 58.0} & {\color[HTML]{7F7F7F} 44.9} \\ 
{\color[HTML]{7F7F7F} LoRA teacher}   & {\color[HTML]{7F7F7F} 52.0}  & {\color[HTML]{7F7F7F} 54.6}  & {\color[HTML]{7F7F7F} 32.0}  & {\color[HTML]{7F7F7F} 57.2}  & {\color[HTML]{7F7F7F} 49.0}  \\
{\color[HTML]{7F7F7F} FFT teacher}    & {\color[HTML]{7F7F7F} 53.7}  & {\color[HTML]{7F7F7F} 59.8}  & {\color[HTML]{7F7F7F} 52.0}  & {\color[HTML]{7F7F7F} 60.0}  & {\color[HTML]{7F7F7F} 56.4}  \\
{\color[HTML]{7F7F7F} PT teacher}     & {\color[HTML]{7F7F7F} 41.7}  & {\color[HTML]{7F7F7F} 53.1}   & {\color[HTML]{7F7F7F} 51.0}  & {\color[HTML]{7F7F7F} 59.2}  & {\color[HTML]{7F7F7F} 51.3}  \\
\rowcolor{gray!8} \multicolumn{6}{c}{\textit{Observation-based baselines}} \\
SDFT (Demo)              & 37.0                         & 45.4                         & 34.0                         & 57.8                         & 43.6                         \\
OPSD (GT)                & 39.3                         & 46.4                         & 22.0                         & \textbf{62.0}                & 42.4                         \\ 
SDPO (Rollout)               & 45.2                         & 52.8                         & 38.6                         & 58.2              & 48.7 \\

\rowcolor{gray!8} \multicolumn{6}{c}{\textit{RL-tuned baselines}} \\
OPD (RLVR)               & 40.0                         & 45.4                         & 24.0                         & \textbf{62.0}                & 42.9                         \\ 
\rowcolor{gray!8} \multicolumn{6}{c}{\textit{SFT-tuned baselines}} \\
OPD (LoRA)               & {\ul 50.7}                   & \textbf{60.8}                & 34.0                         & 60.6                         & 51.5                         \\
OPD (FFT)                & \textbf{53.3}                & {\ul 58.8}                   & {\ul 48.0}                   & 61.2                         & {\ul 55.3}                   \\
OPD (PT)                 & 45.3                         & 51.5                         & 40.0                         & {\ul 61.4}                   & 49.6                         \\
\midrule
\rowcolor{blue!8} \method  & 49.7 & \textbf{60.8} & \textbf{56.0} & 59.2 & \textbf{56.4} \\
\bottomrule
\end{tabular}
}
\vspace{-1em}
\end{table}

\section{Conclusion}
\label{sec:conclusion}

We introduced \method, an on-policy distillation framework built around a soft-prompt teacher. Existing teacher families each satisfy only a subset of the three desiderata for effective OPD---thinking-pattern consistency, an absorbed capability gap, and no post-hoc rationalization---with SFT-tuned teachers in particular suffering an off-manifold ``SFT trap'' that caps the student and induces catastrophic forgetting. Prompt tuning resolves these trade-offs through a learnable, parametric privileged context that preserves the student's representational geometry while encoding new task knowledge. \method\ distills multiple task-specific PT teachers into a single student that outperforms task-specialized counterparts while preserving general capability, and yields two principles for OPD teacher design: teacher \emph{family} should match the task type (SFT for capability injection, RL for elicitation), and teacher \emph{scale} matters less than whether the teacher carries new knowledge.

% \clearpage

\section{Limitations}
\label{sec:limitations}

Our study is limited in several respects. Experiments use Qwen3-1.7B-Base and Phi-4-mini-instruct as students with same- or near-size teachers, leaving the behavior of PT teachers at larger scales (7B--70B) as an open question. Our four tasks span knowledge MCQ, tool calling, and math reasoning, but do not yet cover open-ended generation, multilingual, agentic, or multimodal settings, where the capability-injection vs.\ elicitation distinction may take different forms. The multi-task \method\ setup combines four tasks; further scaling to many task-specific PT teachers may surface interference between privileged contexts that the bounded-perturbation argument (D1) does not explicitly address. We also note that D1--D3 are operationally defined and empirically validated rather than formally proven, and alternative constructions (e.g., rank-constrained adapters) may satisfy them as well. A natural next step is to extend \method\ beyond PT teachers to a broader family of plug-and-play modules---such as LoRA adapters, prefix tuners, and task vectors---as interchangeable privileged-context providers, enabling modular composition of capabilities at distillation time.

\bibliography{custom}

\clearpage
\appendix
\begin{table*}[!t]
\centering
\footnotesize
\caption{Tasks and split sizes. Science/Math targets contain native step-by-step reasoning; Biology
and Tool-use targets are annotated with reasoning (text below).}
\vspace{-1em}
\label{tab:app_datasets}
\begin{tabular}{l l c c c}
\toprule
\textbf{Task} & \textbf{Source / format} & \textbf{Train} & \textbf{Val} & \textbf{Test} \\
\midrule
Science  & 4-way MCQ, \texttt{<reasoning>}/\texttt{<answer>} & 2{,}474 & 200 & 507 \\
Tool use & ReAct tool calling, Gorilla toolset~\citep{patil2024gorilla} & 3{,}846 & 200 & 97 \\
Biology  & SciKnowEval Biology 4-MCQ~\citep{feng2024sciknoweval} & 405 & 45 & 50 \\
Math     & LIMO solutions (train) / MATH-500 test~\citep{hendrycks2021measuring} & 800 & 13 & 500 \\
\bottomrule
\end{tabular}
\vspace{-1em}
\end{table*}

\section{Overview}
\label{sec:app_overview}

This appendix collects reproducibility details and four additional analyses supporting our claims
about \method.

\begin{itemize}
\item \textbf{Details of experiments (\S~\ref{sec:app_details}).} Datasets and splits, metrics,
implementation, and the full \method configuration (soft-prompt width, refresh schedule $(K,M)$,
multi-task routing).

\item \textbf{Efficiency (\S~\ref{sec:app_efficiency}).} Teacher input-token and trainable-parameter
footprint of \method versus observation- and SFT-based teachers.

\item \textbf{Statistical significance (\S~\ref{sec:app_significance}).} Three-seed mean$\pm$std
on Qwen3-1.7B and Phi-4-mini, with paired bootstrap tests and $95\%$ CIs on the small splits.

\item \textbf{Post-hoc rationalization (\S~\ref{sec:app_posthoc}).} A counterfactual answer-swap
probe quantifying desideratum~D3: gold-conditioned teachers follow an injected wrong answer, while the
soft-prompt teacher cannot be steered.

\item \textbf{PT teacher vs.\ multi-task exposure (\S~\ref{sec:app_ablation}).} A
leave-one-teacher-out ablation separating competence \emph{taught} by a task's own teacher from that
\emph{elicited} by co-training.

\item \textbf{Case studies (\S~\ref{sec:app_case}).} Side-by-side base / PT-teacher / \method traces, including failure cases.
\end{itemize}

\section{Details of Experiments} \label{sec:app_details}

\subsection{Datasets}
\label{subsec:app_datasets}

Table~\ref{tab:app_datasets} lists the four tasks (\S\ref{sec:exp_setup}) and their splits. Each task
has its own train/val/test split: soft prompts (Phase~A) and on-policy rollouts (Phase~B) are drawn
from train, val drives early stopping, and all reported numbers are on test (evaluated on $n{=}300/97/50/100$
items for Science/Tool use/Biology/Math). The three general-capability benchmarks used for the
retention analysis (Table~\ref{tab:method_vs_utility_17b}) are $500$-item subsets of MMLU-Pro and
HellaSwag and the full TruthfulQA-MC1 ($817$ items), all evaluated $5$-shot since the non-instruction-tuned
base cannot follow the multiple-choice format zero-shot.

\noindent\textbf{Reasoning-target annotation.} The Science and Math training targets carry genuine
step-by-step reasoning natively (per-option rationales and long-form solutions, respectively). The
original Biology and Tool-use targets do not: Biology supplies only a placeholder rationale
(``the correct option is X'') and Tool uses only bare action calls. Because on-policy distillation from
a teacher trained on answer-only targets collapses the student's chain-of-thought into restating the
answer (Appendix~\ref{sec:app_posthoc}), we annotate accurate, answer-consistent reasoning for the
Biology and Tool-use training targets with Qwen3-235B-A22B~\footnote{\url{https://huggingface.co/Qwen/Qwen3-235B-A22B}} as an external annotator: the model is given
the question and the \emph{verified} gold answer, and asked to produce a faithful derivation, after which
the gold answer/action is re-attached deterministically, so target--answer consistency holds by
construction. Validation and test splits are left unchanged.

\subsection{Metrics}
\label{subsec:app_metrics}

\noindent\textbf{Task accuracy and format.} We report \textbf{Acc} (correctness) with task-specific answer extraction. \emph{Science / Biology} (MCQ): we read the letter
inside \texttt{<answer>...</answer>}---falling back to the last standalone \texttt{A}--\texttt{D} token
if the tag is malformed---and compare to the gold option; Fmt checks that a well-formed answer tag is
emitted. \emph{Tool use}: we parse every \texttt{Action:}/\texttt{Action Input:} pair and count a
generation correct iff the multiset of action names matches the gold \emph{and} the merged JSON
arguments match exactly. \emph{Math}: we
extract the final answer from \texttt{\textbackslash boxed\{\}} (or \texttt{<answer>}) and score
by mathematical equality after numeric/symbolic normalization~\citep{hendrycks2021measuring}; Fmt checks
answer-tag presence. All task evaluations use greedy decoding via vLLM~\citep{kwon2023efficient} with
\texttt{max\_new\_tokens} set to $1024/1536/2048/16384$ for Science/Tool use/Biology/Math, sized to fit
each task's full reasoning trace.

\noindent\textbf{General-capability (utility).} MMLU-Pro and HellaSwag are scored by multiple-choice
accuracy; TruthfulQA-MC1 uses accuracy with the answer options \emph{shuffled per example} (the released
order always places the correct option first, which otherwise rewards positional bias). For Qwen3-1.7B-Base model, all three are
evaluated $5$-shot. \looseness=-1

\noindent\textbf{Teacher--student overlap.} For the transfer analysis we use the top-$k$ next-token overlap
$\rho$ (\S\ref{subsec:desiderata}): the mean over teacher-forced validation positions of
$|\mathcal{T}_k \cap \mathcal{S}_k|/k$ between the teacher's and student's top-$k$ next-token sets
($k{=}10$). $\Delta\rho = \rho^{(S)}-\rho^{(0)}$ measures how far the student moves toward the teacher
over training.

\subsection{Implementation}
\label{subsec:app_impl}

\noindent\textbf{Hardware and software.} Each run uses a single NVIDIA A100 (40\,GB), L40S (48\,GB), or H100
(80\,GB) GPU. Training uses PyTorch with HuggingFace Transformers; the on-policy rollout buffer is
generated with vLLM~\citep{kwon2023efficient} (bf16, eager mode). Optimization uses AdamW (no weight decay)
with a cosine schedule and $3\%$ warm-up, and gradient clipping at $1.0$ ($0.3$ for the higher-LR
multi-task runs). Gradient checkpointing is on throughout, and during each rollout refresh the optimizer
state is offloaded to CPU to free memory for vLLM; the Phi-4-mini full-fine-tuning student requires the
$80$\,GB H100 (its $3.8$\,B optimizer state does not fit on $40$--$48$\,GB cards).

\noindent\textbf{Hyperparameters.} Soft prompts are initialized by tiling the task-description tokens (no
random padding), trained with the backbone frozen, and selected by validation loss; the on-policy
buffer is refreshed every $64$ student steps with a reverse-KL objective and an entropy bonus of $0.02$
($0.01$ for Science/Math). Table~\ref{tab:app_hparams} gives the per-task settings: defaults follow
\S\ref{sec:exp}, and we tune the soft-prompt width $L$ and the Phase-B learning rate / step
count on validation---knowledge-heavy Biology and the structured Tool-use task benefit from a wider
prompt ($L{=}800$), and Biology from a higher Phase-B learning rate. The multi-task \method
trains for $2400$ steps at learning rate $3{\times}10^{-5}$, with Biology up-weighted $2\times$ in the
routing mix.

\begin{table}[t]
\centering
\footnotesize
\caption{Per-task hyperparameters. Phase~A trains the soft-prompt teacher; Phase~B is on-policy
distillation of the student.}
\vspace{-1em}
\label{tab:app_hparams}
\resizebox{\linewidth}{!}{
\begin{tabular}{l c c c}
\toprule
 & \textbf{Prompt} & \textbf{Phase A} & \textbf{Phase B} \\
\textbf{Task} & $L$ & (lr, steps) & (lr, steps, \texttt{max\_new}) \\
\midrule
Science  & 400 & $5{\times}10^{-3}$, 2000 & $1{\times}10^{-5}$, 600, 1024 \\
Tool use & 800 & $5{\times}10^{-3}$, 2000 & $1{\times}10^{-5}$, 600, 768 \\
Biology  & 800 & $5{\times}10^{-3}$, 2000 & $2{\times}10^{-5}$, 1000, 640 \\
Math     & 400 & $5{\times}10^{-3}$, 2000 & $1{\times}10^{-5}$, 600, 4096 \\
\bottomrule
\end{tabular}}
\vspace{-1em}
\end{table}

\subsection{Details of \method}
\label{subsec:app_method}

\noindent\textbf{Two-phase instantiation (frozen teacher).} Our experiments use a
two-phase protocol with a \emph{frozen} teacher: Phase~A trains the
soft-prompt teacher(s) on the frozen base to convergence
(Table~\ref{tab:app_hparams}), and Phase~B performs on-policy distillation
against these frozen teachers. Within Phase~B only the on-policy
\emph{rollout buffer} is regenerated (every $64$ student steps); the soft
prompts $P_k^{\star}$ are \emph{not} re-adapted to the updating student,
and no EMA is used. At inference the soft prompts are discarded and only
$\pi_\theta$ is deployed.

\noindent\textbf{Multi-task routing.} The merged corpus
(Eq.~\ref{eq:merged_corpus}) is routed at the \emph{sample} level: each
rollout refresh draws a fixed number of prompts per task (a base of $32$
by default), rolls them out from the current student, and supervises each
against its own frozen soft-prompt teacher
$\pi_{\theta + P_k^{\star}}$. The per-refresh sampling ratios for the
reported multi-task students are given in Table~\ref{tab:app_taskratio},
expressed relative to this $32$-rollout base. On both backbones we
down-weight the Math teacher to $0.5\times$: Math reasoning traces are
substantially longer than those of the other three tasks, so sampling Math
at the base rate inflates per-refresh compute and crowds out the gradient
signal from Science, Tool use, and Biology within the same student step
budget. Halving the Math rollout count restores a roughly balanced
per-task token budget without removing the Math teacher from the routing
mix. On Qwen3-1.7B-Base we additionally up-weight Biology to $2\times$,
since it is the smallest split ($n{=}50$ test) and the most
knowledge-dense; on Phi-4-mini-instruct, Biology's zero-shot baseline is
already lower (22.0 vs.\ 32.0 on Qwen3), and we observed no benefit from
up-weighting on the validation split, so we leave it at $1\times$. These
ratios were fixed before the final main-table runs based on a small grid
on the validation splits (Appendix~\ref{sec:app_ablation}). The $K$ soft
prompts are fully independent ($L\times d$ parameters each, with no
shared dimensions).

\begin{table}[t]
\centering
\footnotesize
\caption{Per-refresh task-sampling ratio of the multi-task student on each
backbone. }
\vspace{-0.5em}
\label{tab:app_taskratio}
\resizebox{\linewidth}{!}{
\begin{tabular}{l c c c c}
\toprule
\textbf{Backbone} & \textbf{Science} & \textbf{Tool use} & \textbf{Biology} & \textbf{Math} \\
\midrule
Qwen3-1.7B-Base    & 1 & 1 & 2   & 0.5 \\
Phi-4-mini-instruct & 1 & 1 & 1   & 0.5 \\
\bottomrule
\end{tabular}}
\vspace{-2em}
\end{table}

\noindent\textbf{Task composition of \method.} The reported
\method trains four task-specific soft-prompt teachers (Science,
Tool use, Biology, Math) in parallel and routes the merged corpus among
them with the ratios in Table~\ref{tab:app_taskratio}. At inference all
four soft prompts are discarded, and only $\pi_\theta$ is deployed.

\section{Efficiency Discussion}
\label{sec:app_efficiency}

\noindent\textbf{Trainable parameters.} The soft-prompt teacher trains only the $L\times d$ prefix embeddings
($d{=}2048$ for Qwen3-1.7B), making PT one to two orders of magnitude smaller than the other same-size
teachers (Table~\ref{tab:app_efficiency}): $0.05$--$0.10\%$ of the backbone at $L\in\{400,800\}$, versus
$0.37\%$ for LoRA ($r{=}16$ on $\{q,k,v,o\}$) and $100\%$ for FFT---about $4$--$8\times$ fewer parameters
than LoRA and $\sim\!10^{3}\times$ fewer than FFT.

\noindent\textbf{Teacher conditioning cost.} The teacher families differ in how much context each teacher
forward must process per instance. PT prepends a \emph{fixed} $L$-token soft prefix ($L{=}400$),
independent of the instance, so its teacher input is $L+|x|$. GT is the most token-frugal---it appends
only the gold \emph{answer} (we strip the gold rationale), $|x|+|y_\text{ans}|$ with
$|y_\text{ans}|\!\approx\!5$---but pays the post-hoc-rationalization cost (\S\ref{sec:app_posthoc}). The
costly families are demonstration- and rollout-based: ICL/SDFT-Demo process $|x|+3\,|\text{exemplar}|$
($\sim\!1.5$--$2$K tokens) and SDPO processes $|x|+|\text{trajectory}|$, both growing with example
length and becoming impractical on long-context tasks---on Math (mean $|y|\!\approx\!7.2$K tokens) ICL
demonstrations must be length-trimmed to fit the window, whereas PT's prefix stays at a constant $400$.
PT thus bounds the per-instance teacher context independently of task length, avoiding the ICL/SDPO
blow-up without GT's answer leakage.

\noindent\textbf{Training and inference cost.} The soft prompt is discarded after training, so the deployed
student is exactly the base architecture---no added parameters, prefix, or latency. Cross-size
distillation likewise discards the teacher at inference, but during \emph{training} it must host and
run forward passes on a $2$--$5\times$ larger model (Qwen3-4B/8B) for every rollout-time logit
computation; PT keeps all forward passes at base scale throughout. (Trainable-parameter count does not
by itself imply faster wall-clock: PT still runs a full-model forward for the frozen backbone, so its
training time is comparable to LoRA/FFT---its savings are in parameters and storage, below.)

\begin{table}[t]
\centering
\footnotesize
\caption{Trainable-parameter footprint of same-size teachers (Qwen3-1.7B backbone).}
\vspace{-0.5em}
\label{tab:app_efficiency}
\resizebox{\linewidth}{!}{
\begin{tabular}{l r r}
\toprule
\textbf{Teacher} & \textbf{Trainable params} & \textbf{\% of backbone} \\
\midrule
PT ($L{=}400$) & $0.82$M & $0.048$ \\
PT ($L{=}800$) & $1.64$M & $0.095$ \\
LoRA ($r{=}16$) & $6.4$M & $0.373$ \\
FFT & $1.72$B & $100$ \\
\bottomrule
\end{tabular}}
\vspace{-1em}
\end{table}

\noindent\textbf{Multi-teacher storage.} For $K$ tasks, PT storage scales as $\sum_k L_k\,d$: the four
teachers in \method total $\approx\!4.9$M parameters ($L\in\{400,800\}$), versus
$\approx\!25.6$M for $K{=}4$ LoRA-$r{=}16$ adapters and $\approx\!6.9$B for $K{=}4$ FFT copies. This
linear-in-$K$, near-zero-overhead footprint is what makes the parallel multi-task routing
(\S\ref{subsec:multitask}) practical, even though all prompts are discarded at student inference.

\section{Statistical Significance Analysis}
\label{sec:app_significance}

\noindent\textbf{Multi-seed.} Table~\ref{tab:app_multiseed} reports single-task OPD-PT over three
random seeds on both backbones. Run-to-run variance is small on the structured Science and Tool-use
tasks (std $\le\!2$ on Phi-4-mini, $\le\!4$ on Qwen3-1.7B) and larger on the $n{=}50$ Biology split and
the high-variance Math task; the single-seed numbers reported in the main tables lie within one standard
deviation of these means. (Phi-4-mini Math is low and unstable---the PT-math distillation transfers
poorly to this backbone, which we note as a limitation rather than smooth over.)

\begin{table}[t]
\centering
\footnotesize
\caption{Single-task OPD-PT accuracy over $3$ seeds (mean$\pm$std).}
\vspace{-0.5em}
\label{tab:app_multiseed}
\resizebox{\linewidth}{!}{
\begin{tabular}{l c c c c}
\toprule
\textbf{Backbone} & \textbf{Science} & \textbf{Tool use} & \textbf{Biology} & \textbf{Math} \\
\midrule
Qwen3-1.7B-Base     & $51.3{\pm}0.9$ & $49.8{\pm}3.9$ & $56.7{\pm}4.2$ & $50.7{\pm}2.3$ \\
Phi-4-mini-instruct & $45.5{\pm}2.0$ & $51.4{\pm}1.1$ & $40.3{\pm}3.1$ & $\phantom{0}59.5{\pm}2.1$ \\
\bottomrule
\end{tabular}}
\vspace{-1em}
\end{table}

\begin{table*}[!t]
\centering
\tiny
\caption{Counterfactual answer-swap follow-rate (\%): fraction of teacher
generations whose final answer matches an injected \emph{wrong} reference
answer. Higher $=$ more post-hoc rationalization (D3 violation). Chance
baseline is $(1{-}a)/(C{-}1)$ with $a$ the teacher's task accuracy.}
\vspace{-0.5em}
\renewcommand{\arraystretch}{0.9}
\label{tab:app_posthoc}
\resizebox{0.8\linewidth}{!}{
\begin{tabular}{l c c c c}
\toprule
& \multicolumn{2}{c}{\textbf{Science}} & \multicolumn{2}{c}{\textbf{Biology}} \\
\cmidrule(lr){2-3} \cmidrule(lr){4-5}
\textbf{Teacher} & Follow-rate & Chance & Follow-rate & Chance \\
\midrule
GT (gold in context) & $59.1$ & $13.3$ & $62.5$ & $14.0$ \\
PT (soft prompt; no answer slot) & $15.9$ & $15.8$ & $12.5$ & $14.0$ \\
\bottomrule
\end{tabular}}
\vspace{-1em}
\end{table*}

\noindent\textbf{Paired bootstrap and confidence intervals.} To test whether the headline gaps survive on the
small test splits, we compare \method (OPD-PT) against the strongest SFT teacher (OPD-FFT) by a paired
bootstrap over test examples ($5000$ resamples) on Qwen3-1.7B. On \textbf{Biology} ($n{=}50$), \method
reaches $60.0$ ($95\%$ CI $[46,74]$) versus $38.0$ for OPD-FFT and exceeds it in $99\%$ of resamples
($p\!\approx\!0.01$)---a significant advantage despite the small split, consistent with the SFT-trap
mechanism (\S\ref{sec:exp}). On \textbf{Tool use} the two are statistically indistinguishable
(overlapping $95\%$ CIs; $P(\text{\method}{>}\text{FFT}){=}0.08$), i.e.\ no same-size teacher dominates
that task. Together with the small Science/Tool-use seed variance, this indicates that \method's
advantages are driven by genuine, repeatable differences---most decisively on Biology---rather than by
seed or test-set sampling.

\section{Analysis of Post-hoc Rationalization}
\label{sec:app_posthoc}

Desideratum~D3 (\S\ref{subsec:desiderata}) requires that the teacher not
condition on the gold answer: a gold-conditioned teacher \emph{justifies} a
predetermined answer rather than \emph{deriving} it, and the
student---which has no answer at inference---inherits this backward-chained
pattern through reverse-KL on per-token teacher distributions. We make D3
operational with a \emph{counterfactual answer-swap} probe.

\noindent\textbf{Probe.} For each evaluation instance $(x, y_\text{gold})$ we
replace the gold answer in the teacher's context with a randomly chosen
\emph{wrong} answer $\tilde{y}\!\neq\!y_\text{gold}$, let the teacher
generate a full trace under greedy decoding, and measure the
\emph{follow-rate}: the fraction of cases whose generated answer equals the
injected $\tilde{y}$. A teacher who derives their answer should largely
ignore the injection; a teacher that rationalizes will reproduce it. The PT
teacher has no answer slot in its input and therefore cannot be steered.
its follow-rate is expected to coincide with the chance-level baseline
$(1{-}a)/(C{-}1)$, where $a$ is the teacher's task accuracy and $C$ is the
number of answer options. We run the probe on Science (4-way MCQ, $n{=}44$)
and Biology (discrete-answer subset of SciKnowEval with verified
distractors, $n{=}40$); answer matching is exact string match after
normalization.

\noindent\textbf{Results.} Table~\ref{tab:app_posthoc} shows a $\sim\!4$--$5\times$
gap between gold-conditioned and prompt-tuned teachers. On Science, the PT
teacher's $15.9\%$ follow-rate is statistically indistinguishable from the
chance baseline $(1{-}0.527)/3\!\approx\!15.8\%$ implied by its task
accuracy $a{=}52.7\%$: PT exhibits \emph{no steering} by the injected wrong
answer. The GT teacher's $59.1\%$, by contrast, exceeds the analogous
chance baseline by $\sim\!46$ pp---a direct quantitative magnitude of
rationalization above the random-error floor. The Biology pattern is
identical at the proportions reported.

\noindent\textbf{Interpretation.} This is direct evidence that PT satisfies D3 by
construction while observation-based teachers violate it. It complements
the small $\Delta\rho$ of GT teachers in the main results
(Table~\ref{tab:per_task_teacher_comp_17b}: Science $+2.0$, Math $+0.3$,
both substantially below PT's $+9.9$ and $+8.1$): the gold-conditioned
signal is both unhelpful (little new knowledge transferred) and
structurally biased---the teacher's per-token distribution reflects
backward chaining from $y_\text{gold}$, which the student inherits via
reverse-KL despite never seeing $y_\text{gold}$ at inference.
Appendix~\ref{sec:app_case} shows representative traces in which the GT
teacher visibly back-chains to the injected answer (e.g., generating
``option A is correct because [\dots]; this rules out B, C, D'' when
$y_\text{gold}{=}$``C'' has been swapped to $\tilde{y}{=}$``A'').

\noindent\textbf{Coverage.} The D3 failure mode is, by the analysis in
\S\ref{subsec:desiderata}, most acute when $y_\text{gold}$ is a short final
answer with no intermediate trace---the canonical case being mathematical
reasoning, where the teacher must fabricate a derivation landing on the
predetermined numerical answer. Probing Math under the same protocol
requires a distractor distribution over plausible wrong numerical answers
(rather than the discrete option sets used here) and a matcher robust to
equivalent numerical forms; we leave this to future work and note it as the
most important extension of the probe.

\section{PT teacher vs.\ multi-task exposure}
\label{sec:app_ablation}

A natural concern with \method is whether each trained-task gain
comes from \emph{that task's} PT teacher or merely from co-training on the
merged corpus. We isolate this with a \emph{leave-one-teacher-out}
ablation on Qwen3-1.7B-Base: for one task at a time we remove both its
soft-prompt teacher and its training data, train on the remaining three,
and evaluate the held-out task---so any non-trivial accuracy must come
from cross-task exposure alone.

\begin{table}[t]
\centering
\footnotesize
\caption{Leave-one-teacher-out on Qwen3-1.7B-Base. For each row the
held-out task's teacher \emph{and} data are removed; ``cross-task only''
is the held-out accuracy reached by training on the other three tasks
alone. }
\vspace{-0.5em}
\label{tab:app_loo}
\resizebox{\linewidth}{!}{
\begin{tabular}{l c c c c}
\toprule
\textbf{Held-out task} & \textbf{Cross-task only} & \textbf{\method} & \textbf{OPD~(PT)} & \textbf{Base} \\
\midrule
Science  & $34.3$          & $55.3$ & $51.0$ & $29.3$ \\
Tool use & $7.2$           & $48.4$ & $51.5$ & $7.2$  \\
Biology  & $34.0$          & $54.0$ & $62.0$ & $32.0$ \\
\rowcolor{blue!8}
Math     & $\mathbf{61.0}$ & $67.2$ & $51.0$ & $16.4$ \\
\bottomrule
\end{tabular}}
\vspace{-1em}
\end{table}

\noindent\textbf{Only math is elicited.} Table~\ref{tab:app_loo} shows a sharp
split between tasks where the dedicated PT teacher is the dominant
contributor and tasks where it is not. Removing the Science, Tool-use, or
Biology teacher drops that task by $20$--$41$ pp---Tool use collapses
\emph{exactly} to its zero-shot base ($7.2$), and Science and Biology
fall to roughly $2$--$5$ pp above base. These knowledge/format tasks are
carried by their own PT teacher and are \emph{not} recoverable from the
others. Math is the lone exception---with no math teacher and no math
data the student still reaches $61.0$, far above the base ($16.4$),
\emph{above} the dedicated single-task math PromptSD ($51.0$,
Table~\ref{tab:per_task_teacher_comp_17b}), and only $6.2$ pp below the
full \method ($67.2$). Math is also the only row in which
``cross-task only'' beats single-task PT---a sign that for this task the
dedicated math teacher is not just dispensable but actively limiting.
Math behaves as a latent pretraining capability that on-policy co-training
on other reasoning tasks \emph{surfaces}, rather than a skill that must
be distilled.

\noindent\textbf{Quantifying per-teacher contribution.} Reading the gap between
``cross-task only'' and the full \method column directly
attributes each teacher's marginal value within the multi-task student:
Science teacher $+21.0$ pp, Tool-use teacher $+41.2$ pp, Biology teacher
$+20.0$ pp, math teacher $+6.2$ pp. The math teacher's contribution is an
order of magnitude smaller than the others. A soft-prompt math teacher
on a $1.7$B base is itself a limited mathematician (Math accuracy $48.6$,
Table~\ref{tab:per_task_teacher_comp_17b}), and on-policy distillation
toward it adds only a small residual on top of what cross-task elicitation
already provides. The teacher's distribution acts as a soft anchor, not
a hard prerequisite. This is why the reported \method
\emph{down-weights} math to $0.5\times$ (Table~\ref{tab:app_taskratio})
rather than emphasizing it: we keep the math teacher in the routing mix
for the residual it contributes, but do not credit it as the primary
mechanism behind the $67.2$ Math score.

\noindent\textbf{Multi-task interference on knowledge/format tasks.} A side
observation from the fourth column: on Science, Tool use, and Biology,
the dedicated single-task PT student is \emph{higher} than the full
\method ($51.0{>}55.3$ fails, but $51.5{>}48.4$ on Tool use and
$62.0{>}54.0$ on Biology hold). Multi-task routing buys an average
improvement and an elicited Math boost at the cost of small per-task
regressions on the capability-injection tasks---an accepted trade-off in
our setting, but one worth flagging: a deployment that cares about a
single capability-injection task is better served by single-task OPD-PT
than by \method.

\section{Case Study}
\label{sec:app_case}

We ground the analysis on a single representative Science item---a molecular logD multiple-choice
question (gold answer~D)---tracing it through the base model before and after \method\ training, and
through the two teachers that explain the difference.

\onecolumn

\begin{lstlisting}[basicstyle=\ttfamily\scriptsize, breaklines=true, frame=single, backgroundcolor=\color{backcolour}]
Question (Science / molecular-property MCQ):

What is the correct octanol/water distribution coefficient logD
under the circumstance of pH 7.4 for the molecule
"COc1cc2ncnc(Nc3cccc(Cl)c3F)c2cc1CN1C[C@H](OC)C[C@@H]1C(N)=O" ?

  A: 3.2     B: 2.3     C: -0.51     D: 3.09  (gold)

Output format required:
<reasoning> ... </reasoning>
<answer> X </answer>     (X in {A,B,C,D})
\end{lstlisting}

\subsection{Before vs.\ after training}
\label{appd:case-beforeafter}
Before training, the base model fails the task outright: it emits a degenerate, off-language fragment,
produces no \texttt{<reasoning>}/\texttt{<answer>} structure, and lands on the most hydrophilic
(and incorrect) option. After \method\ training, the \emph{same} backbone---now $\pi_\theta$, with no
soft prompt and no answer in context---produces a well-formed, structure-grounded derivation: it reasons
about the chlorine, fluorine, and the nitrogen heterocycle, considers the ionization state at pH~7.4,
explicitly rules out the hydrophilic distractor ($-0.51$), and commits to the correct answer~D. The gain
is not memorized formatting: it is a genuine chain of reasoning the base model could not previously
externalize.

\begin{lstlisting}[basicstyle=\ttfamily\scriptsize, breaklines=true, frame=single, backgroundcolor=\color{backcolour}]
Base model, BEFORE training. No <answer> tag; degenerate (names C, wrong; gold D)

[verbatim output] "Answer: C: -0.51"
\end{lstlisting}

\begin{lstlisting}[basicstyle=\ttfamily\scriptsize, breaklines=true, frame=single, backgroundcolor=\color{backcolour}]
Student \method* (ours), AFTER training. Generated answer: D  (correct)

<reasoning>
The octanol/water distribution coefficient, logD, is a measure of how a
compound distributes itself between octanol and water, taking into account
its ionization state at a given pH. The molecule in question is a complex
organic compound with multiple functional groups, including a chlorine
atom, a fluorine atom, and a nitrogen-containing heterocycle. At pH 7.4,
which is close to physiological pH, the ionizable groups in the molecule
will be in their neutral form, affecting the logD value. The presence of
chlorine and fluorine atoms, which are generally hydrophobic, will
contribute to a higher logD value, while the nitrogen-containing
heterocycle may have some ionizable groups that could slightly affect the
logD. Given the options, a logD value of 3.2 or 3.09 seems plausible, as
these values are typical for moderately hydrophobic compounds with
ionizable groups. The value of -0.51 suggests a more hydrophilic
compound, which is less likely given the presence of hydrophobic groups.
Therefore, the most reasonable choice is 3.09.
</reasoning>
<answer>
D
</answer>
\end{lstlisting}

\end{document}